\documentclass[pdflatex,sn-mathphys-num]{sn-jnl}

\usepackage{graphicx}%
\usepackage{multirow}%
\usepackage{amsmath,amssymb,amsfonts}%
\usepackage{amsthm}%
\usepackage{mathrsfs}%
\usepackage[title]{appendix}%
\usepackage{xcolor}%
\usepackage{textcomp}%
\usepackage{manyfoot}%
\usepackage{booktabs}%
\usepackage{algorithm}%
\usepackage{algorithmicx}%
\usepackage{algpseudocode}%
\usepackage{listings}%

\usepackage{float}
\usepackage{soul}
\usepackage{color}
\usepackage{geometry}
\usepackage{numprint}
\newcommand{\np}[1]{\numprint{#1}}

\colorlet{punct}{red!60!black}
\definecolor{background}{HTML}{EEEEEE}
\definecolor{delim}{RGB}{20,105,176}
\colorlet{numb}{black}
\lstdefinelanguage{json}{
    basicstyle=\normalfont\ttfamily,
    numbers=left,
    numberstyle=\scriptsize,
    stepnumber=1,
    numbersep=8pt,
    showstringspaces=false,
    breaklines=true,
    frame=lines,
    backgroundcolor=\color{background},
    literate=
     *{0}{{{\color{numb}0}}}{1}
      {1}{{{\color{numb}1}}}{1}
      {2}{{{\color{numb}2}}}{1}
      {3}{{{\color{numb}3}}}{1}
      {4}{{{\color{numb}4}}}{1}
      {5}{{{\color{numb}5}}}{1}
      {6}{{{\color{numb}6}}}{1}
      {7}{{{\color{numb}7}}}{1}
      {8}{{{\color{numb}8}}}{1}
      {9}{{{\color{numb}9}}}{1}
      {:}{{{\color{punct}{:}}}}{1}
      {,}{{{\color{punct}{,}}}}{1}
      {\{}{{{\color{delim}{\{}}}}{1}
      {\}}{{{\color{delim}{\}}}}}{1}
      {[}{{{\color{delim}{[}}}}{1}
      {]}{{{\color{delim}{]}}}}{1},
}

\theoremstyle{thmstyleone}%
\theoremstyle{thmstyletwo}%

\theoremstyle{thmstylethree}%

\begin{document}

\title[Article Title]{The Italian lawmaking process: a unified and structured dataset}

\author[1]{\fnm{Matilde} \sur{Contestabile}}\email{matilde.contestabile@santannapisa.it}
\equalcont{These authors contributed equally to this work.}

\author[1]{\fnm{Chiara} \sur{Ferrara}}\email{chiara.ferrara@santannapisa.it}
\equalcont{These authors contributed equally to this work.}

\author[1]{\fnm{Giovanni} \sur{Parrillo}}\email{giovanni.parrillo@santannapisa.it}
\equalcont{These authors contributed equally to this work.}

\author*[1,2]{\fnm{Lorenzo} \sur{Emer}}\email{lorenzo.emer@santannapisa.it}

\author*[1,3]{\fnm{Andrea} \sur{Vandin}}\email{andrea.vandin@santannapisa.it}

\affil*[1]{\orgdiv{Institute of Economics and L'EMbeDS}, 
           \orgname{Scuola Superiore Sant’Anna}, 
           \orgaddress{\street{Piazza Martiri della Libertà, 33}, 
                       \city{Pisa}, \postcode{56127}, 
                       \country{Italy}}}

\affil[2]{\orgdiv{Department of Computer Science}, 
          \orgname{University of Pisa}, 
          \orgaddress{\street{Largo B. Pontecorvo 3}, 
                      \city{Pisa}, \postcode{56126}, 
                      \country{Italy}}}

\affil[3]{\orgdiv{DTU Compute}, 
          \orgname{Technical University of Denmark}, 
          \orgaddress{\street{Anker Engelunds Vej 101}, 
                      \city{Kongens Lyngby}, \postcode{2800}, 
                      \country{Denmark}}}


\abstract{The Italian legislative process is documented in detailed parliamentary records hosted in institutional websites. However, much of this information is available only as unstructured text, limiting its systematic reuse. We applied a controlled, systematic LLM-based extraction procedure, validated against manually annotated records, to convert these texts into ordered and dated events  including, e.g., bill introduction, committee examination, plenary examination, and approval. The resulting dataset reconstructs the procedural histories of all \np{5346} Italian laws enacted between 1985 and 2026, spanning parliamentary terms IX to XIX, for overall \np{101413} events.
Each event records the activity, date, and chamber, with additional procedural details where available. 
The dataset supports research in the emerging fields of computational legal and legislative studies, potentially combined with process-oriented data science methods to examine procedural paths and legislative timing across parliamentary terms.
}




\maketitle

\section*{Background \& Summary}
\label{sec:background}

How bills 
become law is a central question in the study of parliamentary government. In particular, the time a bill spends under examination, the stages through which it passes, and its movement within and between parliamentary chambers provide evidence of how legislative institutions operate: extensive  comparative studies have shown that legislative trajectories are shaped by parliamentary procedures, the organization and powers of chambers, relations among governing parties, and strategic responses to changing political and electoral conditions \cite{Doring1995,BeckerSaalfeld2004,Garwe2022,Imre2023}. Italy provides a particularly informative setting for studying these dynamics because of its symmetric bicameral system, in which the Chamber of Deputies and the Senate exercise equivalent legislative powers and must approve the same text before enactment. As this institutional structure can generate complex legislative trajectories (e.g., involving repeated consideration across chambers), previous research has examined the Italian legislative process in relation to legislative productivity and government coalitions \cite{Kreppel1997,Capano2001}, bicameralism \cite{Zucchini2008}, and the timing and temporal trajectories of legislation \cite{BorghettoGiuliani2012,Borghetto2014}.

Existing resources and databases, including annotated parliamentary debates \cite{erjavec2023parlamint,agnoloni2022parlamintit} and linked collections of speeches, bills, and laws \cite{schwalbach2026parllawspeech}, already provide a basis for such research at the national and international level. For Italy specifically, early work introduced the Italian Law-Making Archive (ILMA), which provides partially structured information on legislation and parliamentary procedures, with coverage limited until 2008 \cite{borghetto2012ilma} \footnote{At the time of writing, the ILMA dataset is no longer available online through its documented web access.}.

A common feature of most existing research is that, while it makes parliamentary texts and legislative information accessible at scale, it still relies on \textit{unstructured} textual form, which requires further extraction and processing before systematic analyses can be performed. 
In particular, in order to conduct quantitative studies of legislation, institutional actions and processes must be identified, dated, and linked to the same law. In Italy, many of these actions are documented in the preparatory work (\textit{lavori preparatori}) accompanying enacted laws\footnote{We retain the English terminology adopted by the official sources of data on the Italian legislation, such as \textit{Normattiva}.}. These records describe the parliamentary history of each act in \textit{narrative form}; extracting their procedural content into a consistent, structured
event-level representation is therefore needed to enable systematic comparison of legislative histories without manually reconstructing each case.


Natural language processing (NLP) offers various methods for this extraction task. Work on legal-domain language models has examined the benefits of adapting models to legal text \cite{chalkidis2020legal}, while LexGLUE and LegalBench provide benchmarks for evaluating legal-language understanding and reasoning \cite{chalkidis2022lexglue,guha2023legalbench}. More generally, large language models (LLMs) and related
architectures have also shown strong performance on complex technical texts, supporting tasks such as classification and information extraction in highly specialized domains \cite{dagdelen2024structured, emer2026patentclassification}. We therefore use an LLM (Gemma 4) to extract legislative events from the narrative preparatory work and assess its accuracy against manually annotated records.

To retrieve the relevant data, we combine two sources: \textit{Normattiva}, Italy’s official portal for accessing enacted national legislation \cite{normattiva}, and parliamentary records published by the Italian Senate \cite{senato_statistiche}, which provide rich metadata. Our dataset comprises two linked components: 1) a \textit{laws registry} containing institutional metadata, and 2) an \textit{event log} recording each law's procedural history. The event log retains both the original text of the preparatory works and the ordered sequence of events extracted from it.
In particular, each event records a procedural activity (e.g., bill introduction, committee assignment, etc.), its date, and the parliamentary chamber involved. The contents of both components, as well as their linking identifiers, are detailed in \textit{Data Records}. 

Our dataset paves the way for further research in the emerging fields of computational legal and legislative studies, as it provides a structured basis for the systematic empirical analysis of lawmaking processes. It also opens up opportunities for applying process mining, i.e., a family of process-oriented data science techniques that reconstruct and analyze processes from recorded events \cite{VanderAalst2016}. These techniques have increasingly been applied to legal and institutional proceedings \cite{caponecchia2025,DBLP:conf/plc2/CaponecchiaDCLV24, Nai2026,contestabile2026bpm,diers2025bundestag,DBLP:conf/bpm/Lopez21}.

\section*{Methods}                   
\begin{figure}
    \centering
    \includegraphics[width=0.99\linewidth]{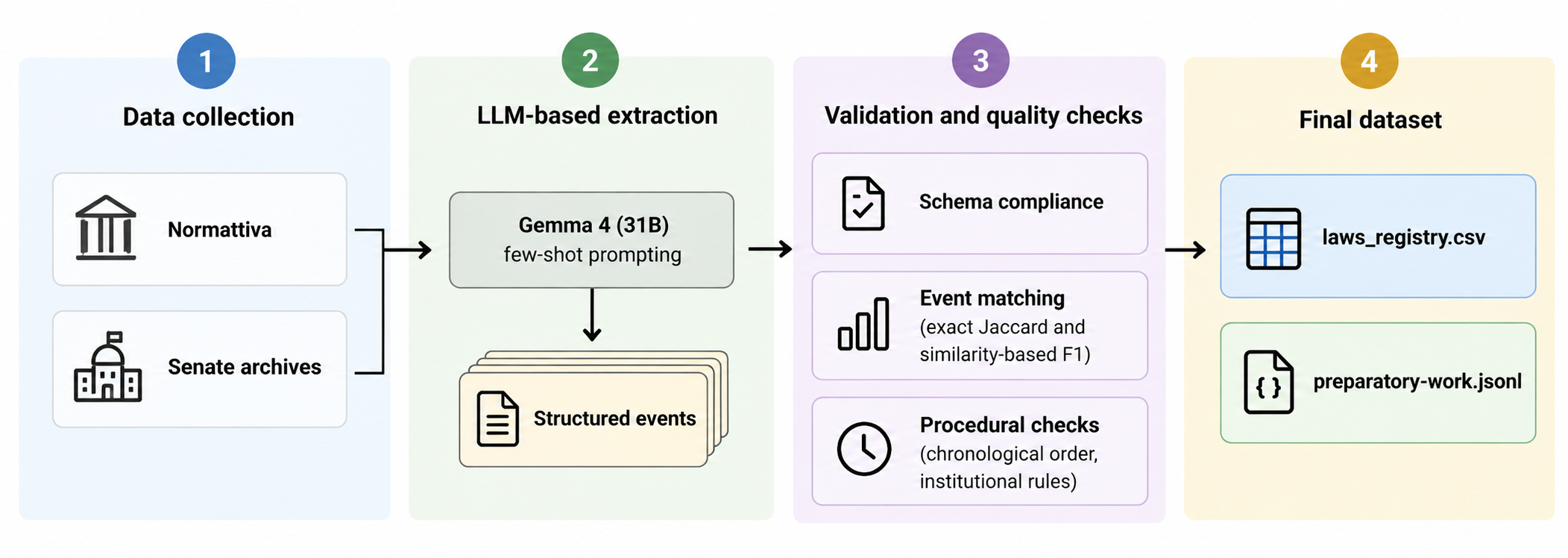}
    \caption{Overview of the adopted methodology.}
    \label{fig:methodology}
\end{figure}

This section describes the construction of the dataset, which combines official legislative records with a systematic extraction pipeline based on an LLM. Specifically, we present the data sources and their acquisition, the extraction process of legislative events from textual preparatory works, the details of the LLM-based pipeline (prompt design, inference configuration), and the post-processing steps applied to the extracted events. A graphical overview of the methodology is displayed in Figure \ref{fig:methodology}.

\subsection*{Data sources and acquisition}

The dataset draws on two official sources: \textit{Normattiva} \cite{normattiva}, and the parliamentary archives of the Italian Senate (\textit{Senato della Repubblica}) \cite{senato_statistiche}. \textit{Normattiva} provides the official texts of enacted laws, their statutory metadata, and the preparatory work (\textit{lavori preparatori}). The Senate archives provide additional information, including law sponsors, type of initiative, and procedural characteristics. 
Notably, as a bill becomes law only after both chambers have approved the same text, the Senate archive provides bicameral coverage, recording legislative activity for both the Senate and the Chamber of Deputies.

Specifically, we collected all \np{5346} laws enacted during Legislatures IX--XIX through \textit{Normattiva}, implementing a web-scraping procedure in compliance with their use terms. For each law, we retrieved its official identifier, title, promulgation date, number of articles, and preparatory work (as defined in \textit{Background \& Summary}). 
We enriched the laws retrieved from \textit{Normattiva} with complementary information from the Senate archives, including attributes identifying the conversion of decree-laws, budget laws, or the occurrence of confidence votes related to the approval of a specific law.
It is important to notice that, while the dataset covers the entire study period, the availability of parliamentary metadata varies across legislatures, reflecting changes in the digitization of institutional records. In particular, more recent laws include richer attributes from the Senate archives, such as legislative initiative, sponsors, confidence votes, and bill consolidation. These differences in metadata availability are thoroughly documented in \textit{Usage Notes}.

\subsubsection*{Event extraction with a large language model}

To extract legislative events from the preparatory work, we used the open-weight \textbf{Gemma 4 (31B)} model \cite{gemmateam2026gemma4}, selected through a comparative evaluation of eight LLMs (see \textit{Technical Validation} for the details). The model was hosted locally on an NVIDIA H200 GPU (141 GB HBM3e memory) using the Ollama serving framework.

We designed an Italian-language prompt to extract legislative events from the preparatory work, preserving the terminology used in parliamentary proceedings. The prompt instructs the model to identify events in the legislative process and return them as structured JSON records, following a predefined schema with ten attributes (Table \ref{tab:event_schema}) \footnote{The complete prompt, including the extraction rules and permitted attribute values, is available in the code repository.}. These attributes include the type of activity, date, parliamentary chamber, committees involved, and other procedural information.

The prompt also specifies rules for handling events. In particular, they must be reported in chronological order, with a sequential identifier (\texttt{sequence\_id}); actions occurring on different dates or involving distinct procedural stages must be recorded as separate events; and attributes must be assigned only when supported by textual evidence.

\begin{table}[!htb]
\centering
\small
\begin{tabular}{llp{9.5cm}}
\toprule
\textbf{Field} & \textbf{Type} & \textbf{Description / Constraints} \\
\midrule
\texttt{sequence\_id} & Integer & Strictly monotonic sequential index ($1, 2, 3, \dots$) \\
\texttt{activity} & String & Controlled vocabulary of 8 stages (\textit{presentation, assignment, committee\_examination, plenary\_examination, approval, report, opinion, severance}) \\
\texttt{date} & String/Null & ISO 8601 date (\texttt{YYYY-MM-DD}) or \texttt{null} \\
\texttt{commissions} & List[String] & Standardized uppercase committee list (e.g., \texttt{["COMMISSIONE 1"]}) \\
\texttt{procedural\_mode} & String/Null & Procedural mode (\textit{referente, legislativa, deliberante, redigente, consultiva}) or \texttt{null} \\
\texttt{sponsors} & List[String] & Primary sponsors or rapporteurs \\
\texttt{has\_co\_sponsors} & Boolean & Flag indicating collective co-sponsorship (``and others'') \\
\texttt{with\_amendments} & Boolean & Flag indicating approval with amendments \\
\texttt{unified\_bills} & List[String] & Identifiers of consolidated bills merged into a unified text \\
\texttt{chamber} & String & Parliamentary chamber (\texttt{"camera"} or \texttt{"senato"}) \\
\bottomrule
\end{tabular}
\caption{Specification of the 10-attribute JSON event extraction schema.}
\label{tab:event_schema}
\end{table}



We adopted a \textit{few-shot prompting} approach, providing the model with three manually annotated examples of legislative histories. These examples were selected to cover different parliamentary procedures, including committee approvals (approval by a parliamentary committee without requiring a plenary vote by the full chamber), unified bills (the consolidation of multiple bills into a single text), bill severance (the division of a bill into separate proposals), and transfers between chambers (the passage of a bill between the Chamber of Deputies and the Senate). The examples were excluded from the reference dataset used for technical validation to avoid overlap. Their identifiers and main procedural characteristics are reported in Table~\ref{tab:fewshot_summary}.

\begin{table}[!htb]
\centering
\small
\begin{tabular}{llcp{9.5cm}}
\toprule
\textbf{Law ID} & \textbf{Origin} & \textbf{Events} & \textbf{Elements included in the examples} \\
\midrule
\texttt{090G0281} & Leg. X & 9 & Decentralized committee passage (\textit{deliberante}), unified texts, and consultative opinions. \\
\texttt{25G00072} & Leg. XIX & 14 & Bill severance (\textit{stralcio}), drafting procedure (\textit{redigente}), and bicameral transfers. \\
\texttt{26G00137} & Leg. XIX & 14 & High committee density ($>10$ bodies), repeated hearings, and bicameral passage. \\
\bottomrule
\end{tabular}
\caption{Overview of the three examples inserted in the prompt (few-shot prompting), including legislature, number of events, and elements providing a rationale for the choice of the examples.}
\label{tab:fewshot_summary}
\end{table}

To ensure complete scientific reproducibility and eliminate stochastic variation across the corpus, inference was performed under strict greedy decoding (temperature $0.0$, seed $42$) with native grammar-constrained JSON schema enforcement (Table \ref{tab:inference_params}). Following standard practice, we disabled the model's \textit{reasoning} mode (\texttt{think: false}) to avoid generating additional reasoning tokens and reduce inference time \cite{Chen2026}. This configuration is appropriate since the task required the extraction of information from source texts into a predefined schema, with no need of intermediate reasoning steps. Finally, to avoid truncation, we examined the token lengths of the preparatory work across all \np{5346} laws and configured a 64k-token context window, which accommodated the prompt, the input texts, and the generated JSON outputs without truncation.

\begin{table}[!htb]
\centering
\small
\begin{tabular}{@{}llp{8.6cm}@{}}
\toprule
\textbf{Parameter} & \textbf{Value} & \textbf{Operational Rationale} \\
\midrule
Model & Gemma 4 (31B) & Open-weight instruction-tuned LLM \\
Hardware & NVIDIA H200 (141~GB) & Local execution, full precision \\
Serving backend & Ollama (JSON mode) & Grammar-constrained decoding for schema validity \\
Decoding temp. & 0.0 & Greedy decoding for deterministic repeatability \\
Random seed & 42 & Fixed seed for exact scientific reproducibility \\
Reasoning / CoT & Disabled (\texttt{think: false}) & Prevents JSON syntax breakage and context bloat \\
Context window & \np{65536} (64k tokens) & Dimensioned against peak corpus length and JSON expansion \\
Max output & \np{49152} (48k tokens) & Guarantees zero sequence truncation on long legislative histories \\
\bottomrule
\end{tabular}
\caption{Inference environment, hardware specifications, and decoding hyperparameters.}
\label{tab:inference_params}
\end{table}

\subsubsection*{Data post-processing}
The raw LLM outputs were post-processed to ensure structural consistency, standardize procedural information, and identify potential irregularities. All outputs were checked for JSON validity and schema compliance. Thanks to grammar-constrained decoding, more than 99.9\% of outputs conformed to the required schema, and no records had to be discarded.

The extracted events were normalized to ensure consistency across Legislatures IX--XIX. Post-processing included five main steps:

\begin{enumerate}
    \item \textbf{Date standardization:} Dates were validated and converted to ISO 8601 format (\texttt{YYYY-MM-DD}). Events spanning multiple dates were separated, and invalid dates were assigned \texttt{null}.
    \item \textbf{Committee normalization:} Committee names were standardized using uppercase labels and Arabic numerals (e.g., \textit{``Commissione I (Affari Costituzionali)''} $\rightarrow$ \texttt{"COMMISSIONE 1"}).
    \item \textbf{Sponsor normalization:} Names were standardized by removing parliamentary titles -- e.g., \textit{``on.''} (Member of Parliament), \textit{``sen.''} (Senator), \textit{``ministro''} (Minister) -- and converting them to uppercase. Collective sponsorship indications -- e.g., \textit{``ed altri''} (``and others'') -- were recorded using the \texttt{has\_co\_sponsors} flag.
    \item \textbf{Unified bill normalization:} References to unified bills were standardized by removing textual prefixes while preserving their official identifiers.
    \item \textbf{Attribute standardization:} Schema keys were translated into English (Table~\ref{tab:event_schema}), and event sequence identifiers (\texttt{sequence\_id}) were reassigned consecutively.
\end{enumerate}
We also implemented a rule-based procedure to identify potential inconsistencies in the extracted event sequences. The checks covered chronological ordering, missing dates, invalid transitions between chambers, and malformed event records. Laws with at least one detected anomaly were flagged using the boolean attribute \texttt{suspicious = true}, rather than excluded from the dataset. This allows researchers to identify potentially problematic records while preserving the full coverage of \np{5346} laws (see \textit{Usage Notes}).

\section*{Data Records}
\label{data_records}

The complete dataset is publicly available on GitHub \cite{prolific_github}. 
The resource is released under the Creative Commons Attribution 4.0 International license (CC BY 4.0). Code and reproducible evaluation pipelines are licensed under the MIT License.

The dataset is organized into two files, covering all \np{5346} Italian laws enacted during Legislatures IX--XIX (1985--2026):

\begin{enumerate}
    \item \texttt{laws\_registry.csv}: A CSV file containing statutory and parliamentary metadata for each law.
    \item \texttt{preparatory\_work.jsonl}: A JSON Lines file containing the original preparatory work and the extracted legislative event sequences, comprising \np{101413} events.
\end{enumerate}

The two files are linked through the unique identifier assigned to each law in the Italian Official Gazette (\textit{Gazzetta Ufficiale}), recorded as \texttt{law\_id} in both files.

\subsection*{Laws Registry (\texttt{laws\_registry.csv})}
The registry captures the institutional trajectory and legislative outcome of all laws, characterizing them with 16 metadata and attributes (Table~\ref{tab:laws_registry_schema}).

    \begin{table}[!htb]
    \centering
    \small
    \resizebox{0.85\textwidth}{!}{%
    \begin{tabular}{llr p{7cm}}
    \toprule
    \textbf{Variable} & \textbf{Type} & \textbf{Availability} & \textbf{Description} \\
    \midrule
    \texttt{law\_id} & String & \np{5346} (100.0\%) & Official Gazette identifier (e.g., \texttt{"26G00136"}), serving as primary key and join   
  key to the procedural event log. \\
    \texttt{law\_number} & String & \np{5346} (100.0\%) & Official statutory law number (e.g., \texttt{"129/2004"}, \texttt{"137/2026"}). \\     
    \texttt{legislature} & Integer & \np{5346} (100.0\%) & Parliamentary legislature index in Arabic numerals (9 to 19). \\
    \texttt{promulgation\_date} & String & \np{3091} (57.8\%) & Official presidential enactment date in ISO 8601 format (\texttt{YYYY-MM-DD}). \\
    \texttt{num\_articles} & Integer & \np{5346} (100.0\%) & Total count of statutory articles composing the enacted statute. \\
    \texttt{initiative} & String & \np{2202} (41.2\%) & Institutional origin (\textit{Government}, \textit{Parliamentary}, \textit{Regional      
  Council}, \textit{Popular}). \\
    \texttt{initiating\_chamber} & String & \np{2192} (41.0\%) & Chamber where the bill was initially introduced (\texttt{"C"}, \texttt{"S"}, or 
  \texttt{"C;S"}). \\
    \texttt{chamber} & String & \np{3107} (58.1\%) & Primary parliamentary chamber of origin (\texttt{"C"} or \texttt{"S"}). \\
    \texttt{sponsors} & String & \np{3106} (58.1\%) & Primary political sponsor(s) or proposing ministers. \\
    \texttt{title} & String & \np{5346} (100.0\%) & Official title and subject matter of the enacted statute. \\
    \texttt{is\_budget\_law} & Boolean & \np{3120} (58.4\%) & Binary indicator for national budget or financial maneuver legislation             
  (Legislatures XIII--XIX). \\
    \texttt{is\_decree\_conversion} & Boolean & \np{3120} (58.4\%) & Binary indicator for conversion of urgent governmental decree-laws          
  (Legislatures XIII--XIX). \\
    \texttt{is\_ratification} & Boolean & \np{3120} (58.4\%) & Binary indicator for international treaty ratification bills (Legislatures XIII-- 
  XIX). \\
    \texttt{confidence\_vote\_granted} & Boolean & \np{3120} (58.4\%) & Indicator of whether a government confidence vote (\textit{fiducia}) was 
  called and passed during parliamentary debate (Legislatures XIII--XIX). \\
    \texttt{is\_merged} & Boolean & \np{2216} (41.5\%) & Indicator of whether the bill was consolidated into a unified text (\textit{testo       
  unificato}; Legislatures XIV--XIX). \\
    \texttt{normattiva\_url} & String & \np{5346} (100.0\%) & Canonical hyperlink to the full statute text on \textit{Normattiva}. \\            
    \bottomrule
    \end{tabular}%
    }
    \caption{Data dictionary for the macro-level Laws Registry (\texttt{laws\_registry.csv}).}
    \label{tab:laws_registry_schema}
    \end{table}

\subsection*{Preparatory Work (\texttt{preparatory\_work.jsonl})}

The event log is provided in JSON Lines format, with one record for each of the \np{5346} laws. Each record contains four attributes:

\begin{itemize}
    \item \texttt{law\_id} (String): Unique identifier of the law.
    \item \texttt{input\_text} (String): Original text of the preparatory work (\textit{lavori preparatori}) retrieved from \textit{Normattiva}.
    \item \texttt{suspicious} (Boolean): Flag indicating whether automated checks detected potential inconsistencies in the extracted event sequence.
    \item \texttt{events} (List[Object]): Chronologically ordered sequence of legislative events, following the ten-attribute schema described in Table \ref{tab:event_schema}.
\end{itemize}

An illustrative example of a structured record from \texttt{preparatory\_work.jsonl} is presented in Listing~\ref{lst:json_sample}.

\begin{lstlisting}[language=json, float=t,caption={Exemplar JSON record representing the procedural trajectory of an enacted statute.}, label={lst:json_sample}, basicstyle=\ttfamily\footnotesize]
{
  "law_id": "087U0041",
  "suspicious": false,
  "input_text": "Camera dei deputati (atto n. 3780): [...]",
  "events": [
    {
      "sequence_id": 1,
      "activity": "presentation",
      "date": "1986-05-21",
      "chamber": "camera",
      "commissions": [],
      "procedural_mode": null,
      "sponsors": ["FALCUCCI"],
      "has_co_sponsors": false,
      "with_amendments": false,
      "unified_bills": []
    },
    [...],
    {
      "sequence_id": 9,
      "activity": "approval",
      "date": "1987-01-29",
      "chamber": "senato",
      "commissions": ["COMMISSIONE 7"],
      "procedural_mode": null,
      "sponsors": [],
      "has_co_sponsors": false,
      "with_amendments": false,
      "unified_bills": []
    }
  ]
}
\end{lstlisting}

\section*{Technical Validation}\label{sec:techVal}

To validate the event extraction procedure, we compare the outputs of eight open-weight LLMs against a manually annotated reference dataset under \textit{zero-shot} and \textit{few-shot} prompting. The reference dataset consists of 100 laws sampled proportionally across the legislatures covered by the corpus (1985--2026). Their preparatory work was manually annotated to identify \np{2031} legislative events.\footnote{The construction of the reference dataset was conducted by three independent annotators. Isolated cases of disagreement were resolved upon discussion; the outputs were then consolidated.}

We evaluate a heterogeneous set of LLMs spanning different model architectures and parameter counts, namely: Qwen~3.5 (9B and 27B) \cite{qwen2026qwen35}, Gemma~4 (12B and 31B) \cite{gemmateam2026gemma4}, Mistral Small~3.2 (24B) \cite{mistral2025small32}, Gemma~3 (27B) \cite{gemmateam2025gemma3}, Qwen3.6-35B-A3B \cite{qwen2026qwen36}, and Llama~3.3 (70B) \cite{meta2024llama33}. Each model receives the extraction instructions either without examples (\textit{zero-shot}) or with three annotated instances (\textit{few-shot}) drawn from outside the reference dataset. All the LLMs we evaluate are open-weight: open-weight models, in fact, offer several advantages over commercial API-based solutions, since they allow local deployment, with greater control over inference settings, ensuring full reproducibility.

We first assess whether the outputs are structurally valid and comply with the event schema. We then compare the event records predicted by each LLM with the reference dataset. In this comparison, exact matching alone can penalize predictions that express the same information as the reference annotations using different wording, as highlighted in recent literature on event argument extraction \cite{fane-etal-2025-bemeae}. We therefore measure both \textit{exact agreement} across all attributes and \textit{approximate agreement} based on semantic similarity.
More in detail, our evaluation comprises three steps:

\begin{enumerate}
    \item \textbf{Schema compliance.}
    First, we check whether outputs form valid JSON and conform to the attribute specification in Table \ref{tab:event_schema} (including field types, date formats, and controlled vocabularies).

    \item \textbf{Exact event matching.} We measure exact agreement between predicted and reference event sets using the Jaccard similarity coefficient \cite{jaccard1901etude}, defined as the size of their intersection divided by the size of their union, \textit{i.e.}, for each law $i$, we compare the predicted event set $\hat{E}_i$ with the reference event set $E_i^*$:
    \begin{equation}
        J_i =
        \frac{|\hat{E}_i \cap E_i^*|}
             {|\hat{E}_i \cup E_i^*|}.
    \end{equation}
    Two events are treated as identical only if all 10 attributes agree. 
    Thus, $J_i=1$ indicates complete agreement, whereas $J_i=0$ indicates that no complete event record matches exactly. 

    \item \textbf{Similarity-based event matching.}
    We complement exact matching with an approximate approach based on semantic similarity, following \cite{tan-etal-2025-cascading}. First, we convert each event record into text using a consistent format. We then encode this text as a vector using \texttt{paraphrase-multilingual-MiniLM-L12-v2} \cite{multilingualminilm}, based on the widely used Sentence-BERT framework \cite{reimers2019sentencebert}. For each law, we compute cosine similarity between predicted and reference event vectors and apply the Hungarian algorithm \cite{kuhn1955hungarian} to identify the one-to-one assignment that maximizes total similarity. We retain assigned pairs with a similarity of at least $\tau=0.80$ as matches, allowing agreement between records that are not identical across all attributes \cite{luo2024zeroshot}.\footnote{We assessed the sensitivity of the results to the similarity threshold by considering alternative values (e.g., $\tau=0.75$ and $\tau=0.85$). The ranking of the evaluated models remained unchanged across these thresholds.} In terms of metrics, we report micro-F1 and macro-F1 scores: micro-F1 aggregates matched and unmatched events across all laws (therefore, assigning more weight to the laws with a higher number of events), whereas macro-F1 averages the F1 scores computed separately for each law, giving equal weight to all laws, independently of the number of events.
\end{enumerate}

Tables \ref{tab:schema_compliance}--\ref{tab:semantic_matching} report the results across the three evaluation measures. Schema compliance was generally high, exceeding 99\% for most model configurations. However, greater differences emerged in exact event matching. Few-shot prompting improved Jaccard similarity for all eight models over zero-shot, with an average increase of 18.3 percentage points. Gemma 4 (31B) achieved the highest exact-match scores in both zero-shot (87.28\%) and few-shot (92.73\%) settings, followed by Qwen 3.5 (27B), which reached 91.29\% in the few-shot configuration. In particular, under few-shot prompting, Gemma~4 (31B) achieved exact agreement ($J=1$) for 58 of the 100 laws, while 89 laws reached a Jaccard score of at least 0.80. These were the highest proportions among the evaluated models, followed by Qwen~3.5 (27B), with 55 and 83 laws, respectively. Gemma~4 (31B) also ranked first when considering the matching approach based on semantic similarity, with micro and macro $F_1$ values of 99.80\% and 99.87\%, respectively, under few-shot prompting.
Based on its performance across the evaluation measures, particularly its higher exact-match agreement, we selected Gemma~4 (31B) with few-shot prompting to extract the events for the complete corpus of \np{5346} laws.

\begin{table}[!htb]
\centering
\small
\begin{tabular}{@{}lccc@{}}
\toprule
\textbf{Model} & \textbf{Params}
& \textbf{Zero-shot (\%)} & \textbf{Few-shot (\%)} \\
\midrule
Gemma 4          & 31B & 99.95 & 99.95 \\
Gemma 4          & 12B & 99.85 & 100.00 \\
Gemma 3          & 27B & 98.95 & 100.00 \\
Qwen 3.5         & 27B & 100.00 & 100.00 \\
Qwen 3.5         & 9B  & 94.55 & 99.25 \\
Qwen 3.6         & 35B & 100.00 & 100.00 \\
Mistral Small 3.2 & 24B & 99.75 & 100.00 \\
Llama 3.3        & 70B & 99.85 & 100.00 \\
\bottomrule
\end{tabular}
\caption{Schema compliance across the eight evaluated
models under zero-shot and few-shot prompting.
Values indicate the reported percentage of outputs
conforming to the required JSON schema.}
\label{tab:schema_compliance}
\end{table}

\begin{table}[!htb]
\centering
\small
\begin{tabular}{@{}lccc@{}}
\toprule
\textbf{Model} & \textbf{Params}
& \textbf{Zero-shot (\%)} & \textbf{Few-shot (\%)} \\
\midrule
\textbf{Gemma 4} & \textbf{31B} &
\textbf{87.28} & \textbf{92.73} \\
Gemma 4          & 12B & 79.53 & 86.30 \\
Gemma 3          & 27B & 62.41 & 89.05 \\
Qwen 3.5         & 27B & 80.00 & 91.29 \\
Qwen 3.5         & 9B  & 25.96 & 72.73 \\
Qwen 3.6         & 35B & 67.41 & 88.90 \\
Mistral Small 3.2 & 24B & 64.96 & 85.02 \\
Llama 3.3        & 70B & 70.26 & 77.84 \\
\bottomrule
\end{tabular}
\caption{Exact-match event-set Jaccard similarity across
the eight evaluated models. Values are percentages
of the reported Jaccard scores, with an event
considered a match only when all ten attributes agree.}
\label{tab:exact_matching}
\end{table}

\begin{table}[!htb]
\centering
\small
\setlength{\tabcolsep}{5pt}
\begin{tabular}{@{}lccccc@{}}
\toprule
& & \multicolumn{2}{c}{\textbf{Zero-shot}}
& \multicolumn{2}{c}{\textbf{Few-shot}} \\
\cmidrule(lr){3-4}\cmidrule(l){5-6}
\textbf{Model} & \textbf{Params}
& \textbf{Micro $F_1$} & \textbf{Macro $F_1$}
& \textbf{Micro $F_1$} & \textbf{Macro $F_1$} \\
\midrule
\textbf{Gemma 4} & \textbf{31B}
& \textbf{99.75} & \textbf{99.77}
& \textbf{99.80} & \textbf{99.87} \\
Gemma 4 & 12B
& 99.53 & 99.43 & 99.28 & 98.82 \\
Gemma 3 & 27B
& 96.34 & 98.41 & 98.23 & 99.18 \\
Qwen 3.5 & 27B
& 99.58 & 99.73 & 99.73 & 99.74 \\
Qwen 3.5 & 9B
& 94.11 & 96.56 & 97.26 & 97.26 \\
Qwen 3.6 & 35B
& 99.10 & 98.76 & 99.48 & 99.51 \\
Mistral Small 3.2 & 24B
& 99.48 & 99.48 & 99.63 & 99.68 \\
Llama 3.3 & 70B
& 98.61 & 99.46 & 98.44 & 99.22 \\
\bottomrule
\end{tabular}
\caption{Similarity-based event matching performance
(micro and macro $F_1$, in percent) under zero-shot
and few-shot prompting. Events were paired using
the Hungarian algorithm with a cosine similarity
threshold of $\tau=0.80$.}
\label{tab:semantic_matching}
\end{table}

\section*{Usage Notes}
\label{sec:usage_notes}

The dataset is provided in CSV and JSON Lines formats and can be analyzed using standard data science tools in Python and R. The two files (\texttt{laws\_registry.csv} and \texttt{preparatory\_work.jsonl}) can be linked through the same law identifier (\texttt{law\_id}). The event sequences can also be converted into formats compatible with process-mining tools such as PM4Py \cite{Berti2019PM4Py}, ProM \cite{VanDongen2005ProM}, and Disco \cite{Gunther2012Disco}. The original preparatory work (\texttt{input\_text}) are retained alongside the extracted events, allowing researchers to inspect the source text and assess individual extractions.

Users should consider several caveats when analyzing the data. First, the dataset includes only enacted laws and therefore does not capture legislative proposals that were rejected, withdrawn, or otherwise failed to become law. Second, although the dataset covers all laws across Legislatures IX--XIX, the availability of metadata varies over time, according to different digitalization practices adopted over time by the Italian Parliament. In particular: for Legislatures IX--XII (1985--1996), records primarily contain statutory information from \textit{Normattiva}; from Legislature XIII (1996--2001), additional attributes become available, including bill identifiers, sponsors, and confidence vote outcomes; and for Legislatures XIV--XIX (2001--2026), the records also include information on legislative initiative, the chamber of introduction, and the consolidation of bills into unified texts. Third, event dates are recorded at daily resolution (\texttt{YYYY-MM-DD}); when multiple events occur on the same date, \texttt{sequence\_id} preserves their reported order but does not provide information about their exact timing. Fourth, coverage is incomplete at both ends of the series: records for Legislature IX (1983–1987) are available only from 1985 onward, while Legislature XIX (beginning in 2022) is covered through June 2026. We nevertheless chose to include both legislatures to provide the broadest temporal coverage possible.

Finally, the dataset includes a quality flag (\texttt{suspicious}) identifying legislative histories with potential inconsistencies detected by automated checks in the post-processing phase. Of the \np{5346} laws, \np{4541} (84.9\%) were not flagged, while 805 (15.1\%) exhibited at least one anomaly. Manual inspection of flagged records revealed that most inconsistencies originated from errors or omissions in the original source text, rather than from the extraction procedure. Researchers can, therefore, use the flag to identify cases requiring further inspection or to conduct robustness checks (e.g., comparing results with and without flagged records).

\section*{Data Availability}         

The complete dataset described in this article is openly available in the GitHub repository at \cite{prolific_github}\footnote{Note to reviewers: Should the paper be accepted, the authors commit to publishing the dataset in a permanent repository (Zenodo) so to adhere to the journal requirements of a public, compliant data repository with a persistent identifier (DOI).
} 
under the \texttt{dataset/} directory. 
The dataset is licensed under the Creative Commons Attribution 4.0 International license (CC BY 4.0).

\section*{Code Availability}         

All source code required to reproduce the dataset construction, few-shot LLM inference, post-processing, anomaly detection, and empirical evaluation is publicly available on GitHub at \cite{prolific_github}. 
The code is implemented in Python (version 3.10+) and released under the open-source MIT License. Complete execution instructions, dependency specifications (\texttt{requirements.txt}), evaluation configurations, and reproduction guidelines are documented in the repository's root and pipeline documentation.

\section*{Author Contributions}

M.C., C.F., and G.P. conceived the study, created and curated the dataset, developed the software pipeline, conducted the experiments and technical evaluation, and drafted the manuscript. L.E. and A.V. contributed to the conceptual design, co-wrote the manuscript, and provided supervision. All authors reviewed and approved the manuscript.

\section*{Competing Interests}
The authors declare no competing interests.


\section*{Funding}
The authors report no external funding for this work.

\bibliography{sn-bibliography}



\end{document}